\documentclass[lnbip]{llncs}
\makeindex          
\usepackage{graphicx}
\usepackage{hyperref}
\usepackage{wrapfig}
\usepackage{colortbl}
\usepackage[USenglish]{babel}
\usepackage{booktabs}
\usepackage[utf8]{inputenc}
\usepackage{soul}
\usepackage{times}
\usepackage{algorithm}
\usepackage[noend]{algpseudocode}
\makeatletter
\def\BState{\State\hskip-\ALG@thistlm}
\makeatother

\usepackage{multirow}
\usepackage{wrapfig,lipsum,booktabs}

\usepackage{amsfonts}
\usepackage{amsmath}
\usepackage{amssymb}
\usepackage{booktabs}
\usepackage{caption}
\usepackage{cite}
\usepackage{comment}
\usepackage{float}
\usepackage[utf8]{inputenc}
\usepackage{subfig}
\usepackage[export]{adjustbox}
\usepackage{todonotes}
\usepackage{wrapfig}
\usepackage{rotating}

\def\post#1{\ensuremath{{#1}\kern-.05ex\bullet}\,}
\usepackage{xcolor}

\usepackage{tikz}
\usetikzlibrary{arrows,backgrounds,calc,decorations.pathmorphing,decorations.pathreplacing,fit,matrix,patterns,petri,positioning, shapes}

\newcommand{\activities}{\mathcal{A}}
\newcommand{\multiset}{\mathcal{B}}

\renewcommand*\abstractname{abstract}
\newcommand\pfun{\mathrel{\ooalign{\hfil$\mapstochar\mkern5mu$\hfil\cr$\to$\cr}}}

\usepackage{diagbox}
\begin{document}
	\mainmatter              

	\title{ Event Log Sampling for Predictive Monitoring}
	\titlerunning{Event Log Sampling for Predictive Monitoring}  

\author{Mohammadreza Fani Sani\inst{1} \and Mozhgan Vazifehdoostirani\inst{2} \and Gyunam Park\inst{1} \and\\ Marco Pegoraro\inst{1} \and Sebastiaan J. van Zelst\inst{3,1} \and Wil M.P. van der Aalst\inst{1,3}}

\authorrunning{Mohammadreza Fani Sani et al.}   
%
%
\institute{$^1$Process and Data Science Chair, 
	RWTH Aachen University, Aachen, Germany\\
	$^2$Industrial Engineering and Innovation Science, Eindhoven University of Technology, Eindhoven, the Netherlands\\ 
	$^3$Fraunhofer FIT, Birlinghoven Castle, Sankt Augustin, Germany \\	
	\email{\{fanisani, gnpark, pegoraro, s.j.v.zelst, wvdaalst\}@pads.rwth-aachen.de}\\
	\email{m.vazifehdoostirani@tue.nl}
	}
	
	\maketitle          

	\begin{abstract}
        Predictive process monitoring is a subfield of process mining that aims to estimate case or event features for running process instances. Such predictions are of significant interest to the process stakeholders. However, state-of-the-art methods for predictive monitoring require the training of complex machine learning models, which is often inefficient. This paper proposes an instance selection procedure that allows sampling training process instances for prediction models. We show that our sampling method allows for a significant increase of training speed for next activity prediction methods while maintaining reliable levels of prediction accuracy.
	\end{abstract}

	\section{Introduction}
	\label{sec:intro}

As the environment surrounding business processes becomes more dynamic and competitive, it becomes imperative to predict process behaviors and take proactive actions~\cite{van_der_aalst_time_2011}. 
Predictive business process monitoring aims at predicting the behavior of business processes, to mitigate the risk resulting from undesired behaviors in the process. 
For instance, by predicting the next activities in the process, one can foresee the undesired execution of activities, thus preventing possible risks resulting from it~\cite{hitfox_group_comprehensible_2016}. 
Moreover, by predicting an expected high service time for an activity, one may bypass or add more resources for the activity~\cite{marquez-chamorro_predictive_2018}. 
Recent breakthroughs in machine learning have enabled the development of effective techniques for predictive business process monitoring. Specifically, techniques based on deep neural networks, e.g., Long-Short Term Memory (LSTM) networks, have shown high performance in 
different tasks~\cite{DBLP:journals/dss/EvermannRF17}. Additionally, the emergence of ensemble learning methods leads to improvement in accuracy in different areas~\cite{breiman1996bagging}. Particularly, for predictive process monitoring, eXtreme Gradient Boosting (XGBoost)~\cite{XGboost} has shown promising results, often outperforming other ensemble methods such as Random Forest or using a single regression tree ~\cite{senderovich2017intra,teinemaa2019outcome}.

Indeed, machine learning algorithms suffer from the expensive computational costs in their training process~\cite{ZHOU2017350}. 
In particular, machine learning algorithms based on neural networks and ensemble learning might require tuning their hyperparameters to be able to provide acceptable accuracy. Such long training time limits the application of the techniques considering the limitations in time and hardware~\cite{DBLP:conf/spaa/PourghassemiZLC20}. This is particularly relevant for predictive business process monitoring techniques. 
Business analysts need to test the efficiency and reliability of their conclusions via repeated training of different prediction models with different parameters~\cite{marquez-chamorro_predictive_2018}. 
Moreover, the dynamic nature of business processes requires new models adapting to new situations in short intervals.

Instance selection aims at reducing original datasets to a manageable volume to perform machine learning tasks, while the quality of the results (e.g., accuracy) is maintained as if the original dataset was used~\cite{10.5555/2671164}.
Instance selection techniques are categorized into two classes based on the way they select instances. 
First, some techniques select the instances at the boundaries of classes. 
For instance, Decremental Reduction Optimization Procedure (DROP)~\cite{10.1023/A:1007626913721} selects instances using \textit{k}-Nearest Neighbors by incrementally discarding an instance if its neighbors are correctly classified without the instance.
The other techniques preserve the instances residing inside classes, e.g., Edited Nearest Neighbor (ENN)~\cite{mnn} preserves instances by repeatedly discarding an instance if it does not belong to the class of the majority of its neighbors.

Such techniques assume independence among instances~\cite{10.1023/A:1007626913721}.
However, in predictive business process monitoring training, instances may be highly correlated~\cite{process-mining}, impeding the application of techniques for instance selection.
Such instances are computed from event data that are recorded by the information system supporting business processes~\cite{de_leoni_general_2016}.
The event data are correlated by the notion of \textit{case}, e.g., patients in a hospital or products in a factory. 
In this regard, we need new techniques for instance selection applicable to event data.

In this work, we suggest an instance selection approach for predicting the next activity, one of the main applications of predictive business process monitoring. By considering the characteristics of the event data, the proposed approach samples event data such that the training speed is improved while the accuracy of the resulting prediction model is maintained.
We have evaluated the proposed methods using two real-life datasets and state-of-the-art techniques for predictive business process monitoring, including LSTM~\cite{LSTM} and XGBoost~\cite{XGboost}.

The remainder is organized as follows. We discuss the related work in \autoref{sec:related}. Next, we present the preliminaries in \autoref{sec:preliminaries} and proposed methods in \autoref{sec:methods}. Afterward, \autoref{sec:evaluation} evaluates the proposed methods using real-life event data and \autoref{sec:discussion} provides discussions. Finally, \autoref{sec:conclusion} concludes the paper.

\section{Related Work}\label{sec:related}
Predictive process monitoring is an exceedingly active field of research. 
At its core, the fundamental component of predictive monitoring is the abstraction technique it uses to obtain a fixed-length representation of the process component subject to the prediction (often, but not always, process traces). In the earlier approaches, the need for such abstraction was overcome through model-aware techniques, employing process models and replay techniques on partial traces to abstract a flat representation of event sequences. Such process models are mostly automatically discovered from a set of available complete traces, and require perfect fitness on training instances (and, seldomly, also on unseen test instances). For instance, van der Aalst et al.~\cite{van_der_aalst_time_2011} proposed a time prediction framework based on replaying partial traces on a transition system, effectively clustering training instances by control-flow information. This framework has later been the basis for a prediction method by Polato et al.~\cite{DBLP:journals/computing/PolatoSBL18}, where the transition system is annotated with an ensemble of SVR and Na{\"i}ve Bayes classifiers, to perform a more accurate time estimation. A related approach, albeit more linked to the simulation domain and based on a Monte Carlo method, is the one proposed by Rogge-Solti and Weske~\cite{DBLP:conf/icsoc/Rogge-SoltiW13}, which maps partial process instances in an enriched Petri net.

Recently, predictive process monitoring started to use a plethora of machine learning approaches, achieving varying degrees of success. For instance, Teinemaa et al.~\cite{teinemaa2016predictive} provided a framework to combine text mining methods with Random Forest and Logistic Regression. Senderovich et al.~\cite{senderovich2017intra} studied the effect of using intra-case and inter-case features in predictive process monitoring and showed a promising result for XGBoost compared to other ensemble and linear methods. A comprehensive benchmark on using classical machine learning approaches for outcome-oriented predictive process monitoring tasks~\cite{teinemaa2019outcome} has shown that the XGBoost is the best-performing classifier among different machine learning approaches such as SVM, Decision Tree, Random Forest, and logistic regression. 

More recent methods are model-unaware and perform based on a single and more complex machine learning model instead of an ensemble. 
The LSTM network model has proven to be particularly effective for predictive monitoring~\cite{DBLP:journals/dss/EvermannRF17,DBLP:conf/caise/TaxVRD17}, since the recurrent architecture can natively support sequences of data of arbitrary length. It allows performing trace prediction while employing a fixed-length event abstraction, which can be based on control-flow alone~\cite{DBLP:journals/dss/EvermannRF17,DBLP:conf/caise/TaxVRD17}, data-aware~\cite{DBLP:conf/ssci/NavarinVPS17}, time-aware~\cite{DBLP:conf/icpm/NguyenCWSME20}, text-aware~\cite{DBLP:conf/bis/PegoraroUGA21}, or model-aware~\cite{DBLP:journals/dss/ParkS20}.

A concept similar to the idea proposed in this paper, and of current interest in the field of machine learning, is \emph{dataset distillation}: utilizing a dataset to obtain a smaller set of training instances that contain the same information (with respect to training a machine learning model)~\cite{wang2020dataset}. While this is not considered sampling, since some instances of the distilled dataset are created ex-novo, it is an approach very similar to the one we illustrate in our paper.
Moreover, recently some instance selection algorithms have been proposed to help process mining algorithms. For example, \cite{sani_2021_sampling,sani_2020_editDistance} proposed to use instance selection techniques to improve the performance of process discovery and conformance checking procedures. 

In this paper, we examine the underexplored topic of event data sampling and selection for predictive process monitoring, with the objective of assessing if and to which extent prediction quality can be retained when we utilize subsets of the training data.

\section{Preliminaries}\label{sec:preliminaries}
In this section, some process mining concepts such as event log and sampling are discussed.
In process mining, we use events to provide insights into the execution of business processes.
Each \textit{event} is related to specific activities of the underlying process. 
Furthermore, we refer to a collection of events related to a specific process instance as a \textit{case}.
Both cases and events may have different attributes. 
An event log that is a collection of events and cases is defined as follows.

\begin{definition}[Event Log]
Let $\mathcal{E}$ be the universe of events, $\mathcal{C}$ be the universe of cases, $\mathcal{AT}$ be the universe of attributes, and $\mathcal{U}$ be the universe of attribute values.
Moreover, let $C{\subseteq}\mathcal{C}$ be a non-empty set of cases, let $E{\subseteq}\mathcal{E}$ be a non-empty set of events, and let $AT{\subseteq} \mathcal{AT}$ be a set of attributes.
We define $(C,E,\pi_C, \pi_E )$ as an event log, where $\pi_C{:}C {\times}\mathcal{AT} {\pfun} \mathcal{U}  $ and $\pi_E{:}E {\times}\mathcal{AT} {\pfun} \mathcal{U} $.
Any event in the event log has a case, therefore, $\nexists_{e\in E} ( \pi_E(e, case) \not\in C)$ and $\bigcup\limits_{e\in E}(\pi_E(e, case)){=} C $.

Furthermore, let $\mathcal{A}{\subseteq}\mathcal{U}$ be the universe of activities and let $\mathcal{V}{\subseteq}\mathcal{A}^*$ be the universe of sequences of activities. 
For  any $e{\in} E$, function $\pi_E(e, activity){\in} \activities$, which means that any event in the event log has an activity. 
Moreover, for any $c{\in}C $ function $\pi_C(c, variant){\in} \mathcal{A}^*{\setminus} \{\langle \rangle\}$ that means any case in the event log has a variant. 
\end{definition}
Therefore, there are some mandatory attributes that are \textit{case} and \textit{activity} for events and \textit{variants} for cases. 
In some process mining applications, e.g., process discovery and conformance checking, just variant information is considered. 
Therefore, event logs are considered as a multiset of sequences of activities. 
In the following, a simple event log is defined. 

\begin{definition}[Simple event log]
Let $\mathcal{A}$ be the universe of activities and let the universe of multisets over a set $X$ be denoted by $\multiset(X)$.
A simple event log is $L{\in} \multiset(\activities^*)  $. 
Moreover, let $\mathcal{EL} $ be the universe of event logs and $EL{=}(C,E,\pi_C,\pi_E){\in} \mathcal{EL} $ be an event log.
We define function $sl{:}\mathcal{EL}{\to} \multiset(\{ \pi_E(e,activity) | e{\in}E  \}^*)$ returns the simple event log of an event log.
The set of unique variants in the event log is denoted by $\overline{sl(EL)}$.
\end{definition}
Therefore, $sl$ returns the multiset of variants in the event logs. 
Note that the size of a simple event log equals the number of cases in the event logs, i.e., $ sl(EL){=}|C|$

In this paper, we use sampling techniques to reduce the size of event logs. 
An event log sampling method is defined as follows. 

\begin{definition}[Event log sampling]
Let $\mathcal{EL}$ be the universe of event logs and $\activities$ be the universe of activities. 
Moreover, let $EL{=}(C,E,\pi_C,\pi_E){\in} \mathcal{EL} $ be an event log, we define function $\delta{:}\mathcal{EL}{\to} \mathcal{EL}  $ that returns the sampled event log where if $(C',E',\pi'_C, \pi'_E){=}\delta (EL)$, then $C'{\subseteq}C$, $E'{\subseteq}E$, $\pi'_e{\subseteq}\pi_E$, $\pi'_C{\subseteq}\pi_C$, and consequently, $\overline{sl(\delta(EL))} {\subseteq} \overline{sl(EL)}$. 
We define that $\delta$ is a variant-preserving sampling if $\overline{sl(\delta(EL))} {=} \overline{sl(EL)}$.
\end{definition}
In other words, a sampling method is variant-preserving if and only if all the variants of the original event log are presented in the sampled event log.

To use machine learning methods for prediction, we usually need to transfer each case to one or more features. 
The feature is defined as follows. 

\begin{definition} [Feature]
Let $\mathcal{AT}$ be the universe of attributes, $\mathcal{U}$ be the universe of attribute values, and $\mathcal{C}$ be the universe of cases. 
Moreover, let $AT{\subseteq}\mathcal{AT}$ be a set of attributes. 
A feature is a relation between a sequence of attributes' values for $AT$ and the target attribute value, i.e., $f{\in} (\mathcal{U}^{|AT|} {\times} \mathcal{U} ) $.
We define $\mathit{fe}{:}\mathcal{C}{\times}\mathcal{EL}{\to}\multiset(\mathcal{U}^{|AT|} {\times} \mathcal{U} )$ is a function that receives a case and an event log, and returns a multiset of features. 
\end{definition}
For the next activity prediction, i.e., our prediction goal, the target attribute value should be an activity.
Moreover, a case in the event log may have different features. 
For example, suppose that we only consider the activities.
For the case $\langle a,b,c,d \rangle$, we may have $(\langle a \rangle, b)$, $(\langle a,b \rangle, c)$, and $(\langle a,b,c \rangle, d)$ as features. 
Furthermore, $\sum\limits_{c\in C} fe(c,EL)$ are the corresponding features of event log $EL{=}(C,E,\pi_C,\pi_E)$ that could be given to different machine learning algorithms. 
For more details on how to extract features from event logs please refer to \cite{qafari2020feature}.

\section{Proposed Sampling Methods}\label{sec:methods}
\begin{figure}[tb]
    \centering
    \includegraphics[width=1.01\textwidth]{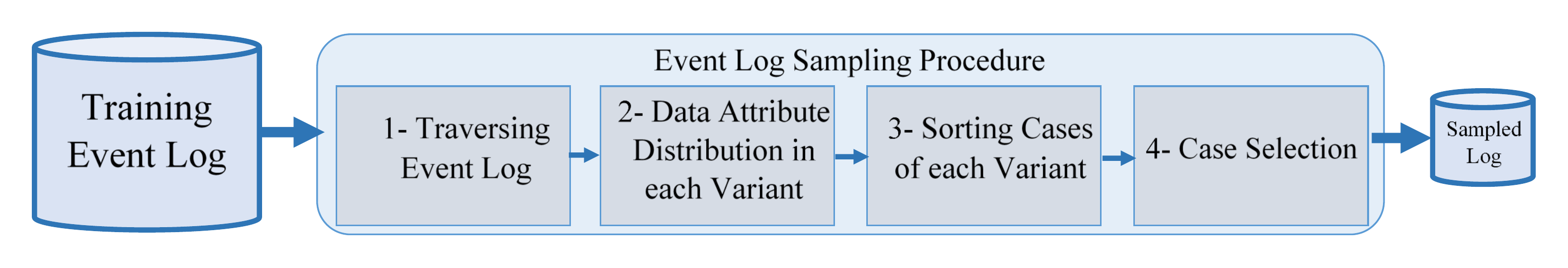}
    \caption{A schematic view of the proposed sampling procedure}
    \label{fig:samplingProcedure}
\end{figure}

In this section, we propose an event log preprocessing procedure that helps prediction algorithms to perform faster while maintaining reasonable accuracy. 
The schematic view of the proposed sampling approach is presented in \autoref{fig:samplingProcedure}.
We first need to traverse the event log and find the variants and corresponding traces of each variant in the event log. 
Moreover, different distributions of data attributes in each variant will be computed.  
Afterward, using different sorting and instance selection strategies, we are able to select some of the cases and return the sample event log. 
In the following, each of these steps is explained in more detail. 

\begin{enumerate}
    \item \textit{Traversing the event log}: In this step, the unique variants of the event log and the corresponding traces of each variant are determined.
    In other words, consider event log $EL$ that $\overline{sl(EL)}{=}\{\sigma_1, ...,\sigma_n \}$ where $n{=}|\overline{sl(EL)}|$, we aim to split $EL$ to $EL_1, .., EL_{n}$ where $EL_i$ only contains all the cases that $C_i{=}\{c{\in}C | \pi_C(c,variant){=} \sigma_i  \}$ and $E_i{=}\{e{\in}E | \pi_E(e,case){\in}C_i \}$. 
    Obviously, $\bigcup\limits_{1\leq i\leq n }(C_i){=}C$ and $\bigcap\limits_{1\leq i\leq n }(C_i){=}\varnothing$. 
    \item \textit{Distribution Computation}: In this step, for each variant of the event log, we compute the distribution of different data attributes $a{\in} AT$. 
    It would be more practical if the interesting attributes are chosen by an expert. Both event and case attributes can be considered. 
    A simple approach is to compute the frequency of categorical data values. 
    For numerical data attributes, it is possible to consider the average or the median of values for all cases of each variant. 
    
    \item \textit{Sorting the cases of each variant}: In this step, we aim to sort the traces of each variant. 
    We need to sort the traces to give a higher priority to those traces that can represent the variant better. 
    One way is to sort the traces based on the frequency of the existence of the most occurred data values of the variant. 
    For example, we can give a higher priority to the traces that have more frequent resources of each variant.
    It is also possible to sort the traces based on their arrival time or randomly.    
    \item \textit{Returning sample event logs}: Finally, depending on the setting of the sampling function, we return some of the traces with the highest priority for all variants.
    The most important point about this step is to know how many traces of each variant should be selected. 
    In the following, some possibilities will be introduced. 
    \begin{itemize}
        \item \textit{Unique selection}: In this approach, we select only one trace with the highest priority. In other words, suppose that $L'{=}sl(\delta(EL))$, $\forall_{\sigma\in L'} L'(\sigma){=}1 $.
        Therefore, using this approach we will have $|sl(\delta(EL))|{=} |\overline{sl(EL)}|$. It is expected that using this approach, the distribution of frequency of variants will be changed and consequently the resulted prediction model will be less accurate.  
        \item \textit{Logarithmic distribution}: In this approach, we reduce the number of traces in each variant in a logarithmic way. If $L{=}sl(EL)$ and $L'{=}sl(\delta(EL))$, $\forall_{\sigma\in L'} L'(\sigma){=}{[}Log_{k}(L(\sigma)) {]} $.
        Using this approach, the infrequent variants will not have any trace in the sampled event log.  
        By using a higher $k$, the size of the sampled event log is reduced more. 
        \item \textit{Division}: This approach performs similar to the previous one, however, instead of using logarithmic scale, we apply the division operator. In this approach, $\forall_{\sigma\in L'} L'(\sigma){=}{\lceil}\frac{(\sigma)}{k} {\rceil} $.
        A higher $k$ results in fewer cases in the sample event log. 
        Note that using this approach all the variants have at least one trace in the sampled event log.
    \end{itemize}
    There is also a possibility to consider other selection methods. For example, we can select the traces completely randomly from the original event log.
\end{enumerate}

By choosing different data attributes in Step 2 and different sorting algorithms in Step 3, we are able to lead the sampling of the method on \textit{which} cases should be chosen. 
Moreover, by choosing the type of distribution in Step 4, we determine \textit{how many} cases should be chosen.
To compute how sampling method $\delta$ reduces the size of the given event log $EL$, we use the following equation:
\begin{equation}
    R_{S}{=}\frac{|sl(EL)|}{|sl(\delta(EL))|}
\end{equation}
The higher $R_{S}$ value means, the sampling method reduces more the size of the training log. 
By choosing different distribution methods and different \textit{k-values}, we are able to control the size of the sampled event log.
It should be noted that the proposed method will apply just to the training event log. In other words, we do not sample event logs for development and test datasets.  
\vspace{-0.2cm}
\section{Evaluation}\label{sec:evaluation}
In this section, we aim at designing some experiments to answer our research question, i.e., "Can we improve the computational performance of prediction methods by using the sampled event logs, while maintaining a similar accuracy?".
It should be noted that the focus of the experiments is not on prediction model tuning to have higher accuracy. Conversely, we aim to analyze the effect of using sampled event logs (instead of the whole datasets) on the required time and the accuracy of prediction models.  
In the following, we first explain the event logs that are used in the experiments. 
Afterward, we provide some information about the implementation of sampling methods. 
Moreover, the experimental setting is discussed and, finally, we show the experimental results. 
\vspace{-0.2cm}
\subsection{Event logs}
To evaluate the proposed sampling procedure for prediction, we have used two event logs widely used in the literature.
Some information about these event logs is presented in \autoref{tab:eventLogs}.
In the \textit{RTFM} event log, which corresponds to a road traffic management system, we have some high frequent variants and several infrequent variants. Moreover, the number of activities in this event log is high. 
Some of these activities are infrequent, which makes this event log imbalanced.
In the \textit{BPIC-2012-W} event log, relating to a process of an insurance company, the average of variant frequencies is lower. 

\begin{table}[tb]
\caption{Overview of the event logs that are used in the experiments. The accuracy and the required times (in seconds) of different prediction methods for these event logs are also presented.\label{tab:eventLogs} }
\scriptsize
\begin{adjustbox}{width=\columnwidth, center}
\begin{tabular}{|l|l|l|l|l|l|l|l|l|l|}
\hline
Event Log & Cases& Activities & Variants & Attributes & FE Time & \textit{LSTM} Train Time & \textit{LSTM} \textit{Acc} & \textit{XG} Train Time & \textit{XG Acc} \\ \hline
\textit{RTFM} & 150370 & 11 & 231 & 1 & 73649 & 3021 & 0.791 & 11372 & 0.814 \\ \hline
\textit{BPIC-2012-W} & 9658 & 6 & 2643 & 2 & 1212 & 3344 & 0.68 & 2011 & 0.685 \\ \hline
\end{tabular}
\end{adjustbox}
\normalsize
\vspace{-0.4cm}
\end{table}
\vspace{-0.2cm}
\subsection{Implementation}
We have developed the sampling methods as a plug-in in the ProM framework~\cite{Prom}, accessible via \url{https://svn.win.tue.nl/repos/prom/Packages/LogFiltering}.
This plug-in takes an event log and returns k different train and test event logs in the CSV format. 
Moreover, to train the prediction method, we have used XGBoost~\cite{XGboost} and LSTM~\cite{LSTM} methods as they are widely used in the literature and outperformed their counterparts. 
Our LSTM network consisted of an input layer, two LSTM layers with \textit{dropout} rates of $10\%$, and a dense output layer with the \textit{SoftMax} activation function. We used “categorical cross-entropy” to calculate the loss and adopted \textit{ADAM} as an optimizer. 
We used \textit{gbtree} with a max depth of $6$ as a booster in our XGBoost model. Uniform distribution is used as the sampling method inside our XGBoost model. To avoid overfitting in both models, the training set is further divided into $90\%$ training set and $10\%$ validation set to stop training once the model performance on the validation set stops improving. We used the same setting of both models for original event logs and sampled event logs. 
To access our implementations of these methods and the feature generation please refer to \url{https://github.com/gyunamister/pm-prediction/}.
For details of the feature generation and feature encoding steps, please refer to \cite{DBLP:journals/dss/ParkS20}.
\vspace{-0.2cm}
\subsection{Evaluation setting}
To sample the event logs, we use three distributions that are \textit{$log$ distribution}, \textit{division}, and \textit{unique variants}. 
For the $log$ distribution method, we have used $2,3$, and $10$ (i.e., $log_2, log_3$, and $log_{10}$).
For the division method, we have used $2,5$, and $10$ (i.e., $d2, d5$, and $d10$). 
For each event log and for each sampling method, we have used a $5\text{-}fold$ cross-validation. 
Moreover, as the results of the experiments are non-deterministic, all the experiments have been repeated $5$ times and the average values are represented.

Note that, for both training and evaluation phases, we have used the same settings for extracting features and training prediction models. We used one-hot encoding to encode the sequence of activities for both LSTM and XGBoost models. 
We ran the experiment on a server with Intel Xeon CPU E7-4850 2.30GHz, and 512 GB of RAM. 
In all the steps, one CPU thread has been used. 
We employed the \textit{Weighted Accuracy} metric~\cite{Accuracy} to compute how a prediction method performs for test data. 
To compare the accuracy of the prediction methods, we use the \textit{relative accuracy} that is defined as follows. 
\vspace{-0.2cm}
\begin{equation}
     R_{Acc}{=}\frac{\text{ Accuracy using the sampled training log}}{\text{Accuracy using the whole training log}}
\end{equation}
If $R_{Acc}$ is close to $1$, it means that using the sampling event logs, the prediction methods behave almost similar to the case that the whole data is used for the training. Moreover, values higher than $1$ indicate the accuracy of prediction methods has improved. 

To compute the improvement in the performance of training time, we will use the following equations. 
\vspace{-0.2cm}
\begin{equation}
    R_{t}{=}\frac{\text{Training time using whole data}}{\text{Training time using the sampled data}}
\end{equation}
\vspace{-0.2cm}
\begin{equation}
    R_{FE}{=}\frac{\text{Feature extraction time using whole data}}{\text{Feature extraction time using the sampled data}}
\end{equation}
For both equations, the resulting values indicate how many times the sampled log is faster than using all data. 
\vspace{-0.2cm}
\subsection{Experimental results}
\autoref{tab:R&PIFE} presents the reduction rate and the improvement in the feature extraction phase using different sampling methods.
As it is expected, the highest reduction rate is for \emph{$log_{10}$} (as it removes infrequent variants and keeps few traces of frequent variants), and respectively it has the biggest improvement in $R_{FE}$. 
Moreover, the lowest reduction is for \emph{d2}, especially if there are lots of unique variants in the event log (i.e., for the \textit{RTFM} event log).  
We expected smaller event logs to require less feature extraction time. However, results indicate that the relationship is not linear, and by having more reduction in the size of the sampled event log there will be a much higher reduction in the feature extraction time. 

In \autoref{tab:LSTM} and \autoref{tab:XG}, the results of improvement in $R_{t}$ and $R_{Acc}$ are shown for LSTM and XG prediction methods. 
As expected, by using fewer cases in the training, the performance of training time improvement will be higher. 
Comparing the results in these two tables and the results in \autoref{tab:R&PIFE}, it is interesting to see that in some cases, even by having a high reduction rate, the accuracy of the trained prediction model is close to the case in which whole training log is used. 
For example, using $d{10}$ for the \textit{RTFM} event log, we will have high accuracy for both prediction methods.
In other words, we are able to improve the performance of the prediction procedure while the accuracy is still reasonable. 

When using the LSTM prediction method for the RTFM event log, there are some cases where we have accuracy improvement. 
For example, using $d{3}$, there is a $0.4\%$ improvement in the accuracy of the trained model.
It is mainly because of the existence of high frequent variants. 
These variants lead to having unbiased training logs and consequently, the accuracy of the trained model will be lower for infrequent behaviors.

\begin{table}[t]
\caption{The reduction in the size of training logs (i.e., $R_{S}$) and the improvement in the performance of feature extraction part (i.e., $R_{FE}$) using different sampling methods.\label{tab:R&PIFE} }\scriptsize
\begin{adjustbox}{width=\columnwidth, center}
\begin{tabular}{|l|c|c|c|c|c|c|c|c|c|c|c|c|c|c|}
\hline
 Sampling Methods& \multicolumn{2}{c|}{d2} & \multicolumn{2}{c|}{d3} & \multicolumn{2}{c|}{d10} & \multicolumn{2}{c|}{$log_{2}$} & \multicolumn{2}{c|}{$log_{3}$} & \multicolumn{2}{c|}{$log_{10}$} & \multicolumn{2}{c|}{unique} \\ \hline
 Event Log& $R_{S}$ & $R_{FE}$ & $R_{S}$ & $R_{FE}$ & $R_{S}$ & $R_{FE}$ & $R_{S}$ & $R_{FE}$ & $R_{S}$ & $R_{FE}$ & $R_{S}$ & $R_{FE}$ & $R_{S}$ & $R_{FE}$ \\ \hline
RTFM\cite{de_2015_road} & 1.99 & 4.8 & 3.0 & 11.1 & 9.8& 106.9 & 153.5 & 12527.6 &236.3 & 23699.2 & \textbf{572.3} & \textbf{74912.8} & 285.1& 24841.8 \\ \hline
BPIC-2012-W~\cite{BPI_2012_challeng} & 1.22 & 1.37 & 1.41 & 1.80 & 1.66
 & 2.51 & 6.06 & 22.41 & 9.05 & 37.67 & 28.50 & 208.32 & 1.73 & 2.36 \\ \hline
\end{tabular}
\end{adjustbox}

\caption{The accuracy and the improvement in the performance of prediction using different sampling methods for LSTM.\label{tab:LSTM} }
\begin{adjustbox}{width=\columnwidth, center}
\begin{tabular}{|l|c|c|c|c|c|c|c|c|c|c|c|c|c|c|}
\hline
 Sampling Methods& \multicolumn{2}{c|}{d2} & \multicolumn{2}{c|}{d3} & \multicolumn{2}{c|}{d10} & \multicolumn{2}{c|}{$log_{2}$} & \multicolumn{2}{c|}{$log_{3}$} & \multicolumn{2}{c|}{$log_{10}$} & \multicolumn{2}{c|}{unique} \\ \hline
 Event Log& $R_{Acc}$ & $R_{t}$ & $R_{Acc}$ & $R_{t}$ & $R_{Acc}$ & $R_{t}$ & $R_{Acc}$ & $R_{t}$ & $R_{Acc}$ & $R_{t}$ & $R_{Acc}$ & $R_{t}$ & $R_{Acc}$ & $R_{t}$ \\ \hline
RTFM & 1.001 & 2.0 & \textbf{1.004} & 2.9 & 0.990& 9.0  & 0.716 & 26.7 & 0.724 & 33.0 & 0.767 & 41.8 &  0.631 & 29.1 \\ \hline
BPIC-2012-W &1.000 & 1.4 & 0.985 & 1.3 & 0.938 & 1.3 & 0.977 & 4.7 & 0.970 & 5.8 & 0.876 & 11.9 & 0.996 & 1.6 \\ \hline
\end{tabular}
\end{adjustbox}

\caption{The accuracy and the improvement in the performance of prediction using different sampling methods for XGBoost.\label{tab:XG} }
\begin{adjustbox}{width=\columnwidth, center}
\begin{tabular}{|l|c|c|c|c|c|c|c|c|c|c|c|c|c|c|}
\hline
 Sampling Methods& \multicolumn{2}{c|}{d2} & \multicolumn{2}{c|}{d3} & \multicolumn{2}{c|}{d10} & \multicolumn{2}{c|}{$log_{2}$} & \multicolumn{2}{c|}{$log_{3}$} & \multicolumn{2}{c|}{$log_{10}$} & \multicolumn{2}{c|}{unique} \\ \hline
 Event Log& $R_{Acc}$ & $R_{t}$ & $R_{Acc}$ & $R_{t}$ & $R_{Acc}$ & $R_{t}$ & $R_{Acc}$ & $R_{t}$ & $R_{Acc}$ & $R_{t}$ & $R_{Acc}$ & $R_{t}$ & $R_{Acc}$ & $R_{t}$ \\ \hline
RTFM & \textbf{1.000} & 2.4 & \textbf{1.000} & 1.4 & \textbf{1.000} & \textbf{84.1}  & 0.686 & 126.4 & 0.706 &191.8 & 0.772 &355.0 & 0.582 & 297.7 \\ \hline
BPIC-2012-W & 0.999 & 2.3 & 0.998 & 2.4 & 0.997 & 3.4 & 0.923 & 10.7 & 0.970 & 16.7 & 0.883 & 64.8 & 0.997 & 2.8 \\ \hline
\end{tabular}
\end{adjustbox}
\vspace{-0.4cm}

\end{table}

\vspace{-0.3cm}
\section{Discussion}\label{sec:discussion}
The results indicate that we do not always have a typical trade-off between the accuracy of the trained model and the performance of the prediction procedure. 
In other words, there are some cases where the training process is much faster than the normal procedure, even though the trained model provides an almost similar accuracy. 
We did not provide the results for other metrics; however, there are similar patterns for weighted recall, precision, and f1-score. 
Thus, the proposed sampling methods can be used when we aim to apply hyperparameter optimization~\cite{Hyperparameter}. 
In this way, more settings can be analyzed in a limited time. Moreover, it is reasonable to use the proposed method when we aim to train an online prediction method or on naive hardware such as cell phones.

Another important outcome of the results is that for different event logs, we should use different sampling methods to achieve the highest performance. 
For example, for the \textit{RTFM} event log---as there are some highly frequent variants---the division distribution may be more useful. In other words, independently of the used prediction method, if we change the distribution of variants (e.g., using $unique$ distribution), it is expected that the accuracy will sharply decrease.
However, for event logs with a more uniform distribution, we can use logarithmic and unique distributions to sample event logs.
The results indicate that the effect of the chosen distribution (i.e., $unique$, $division$, and $logarithmic$) is more important than the used \textit{k-value}. 
Therefore, it would be valuable to investigate more on the characteristics of the given event log and suitable sampling parameters for such distribution.
For example, if most variants of a given event log are unique, the $division$ and $unique$ methods are not able to have remarkable $R_{S}$ and consequently, $R_{FE}$ and $R_{t}$ will be close to $1$.

Moreover, results have shown that by oversampling the event logs, although we will have a very big improvement in the performance of the prediction procedure, the accuracy of the trained model is significantly lower than the accuracy of the model that is trained by the whole event log. 
Therefore, we suggest gradually increasing (or decreasing) the size of the sampled event log in the hyper-parameter optimization scenarios. 

By analysis of the results using common prediction methods, we have found that the infrequent activities can be ignored using some hyper-parameter settings. 
This is mainly because the event logs are unbalanced for these infrequent activities.
Using the sampling methods that modify the distribution of the event logs such as the $unique$ method can help the prediction methods to also consider these activities.

Finally, in real scenarios, the process can change because of different reasons~\cite{josep_2012_drift}. 
This phenomenon is usually called \textit{concept drift}. 
By considering the whole event log for training the prediction model, it is most probable that these changes are not considered in the prediction. Using the proposed sampling procedure, and giving higher priorities to newer traces, we are able to adapt to the changes faster, which may be critical for specific applications.  

\vspace{-0.4cm}
\section{Conclusion}\label{sec:conclusion}
\vspace{-0.1cm}
In this paper, we proposed to use the subset of event logs to train prediction models. 
We proposed different sampling methods for next activity prediction. 
These methods are implemented in the ProM framework. 
To evaluate the proposed methods, we have applied them on two real event logs and have used two state-of-the-art prediction methods: LSTM and XGBoost. 
The experimental results have shown that, using the proposed method, we are able to improve the performance of the next activity prediction procedure while retaining an acceptable accuracy (in some experiments, the accuracy increased).
However, there is a relation between event logs characteristics and suitable parameters that can be used to sample these event logs. 
The proposed methods can be helpful in situations where we aim to train the model fastly or in hyper-parameter optimization scenarios. 
Moreover, in cases where the process can change over time, we are able to adapt to the modified process more quickly using sampling methods. 

To continue this research, we aim to extend the experiments to study the relationship between the event log characteristics and the sampling parameters. 
Additionally, we plan to provide some sampling methods that help prediction methods to predict infrequent activities, which could be more critical in the process.
Finally, it is interesting to investigate more on using sampling methods for other prediction method applications such as last activity and remaining time prediction.
\vspace{-0.3cm}
\section*{Acknowledgment}
\vspace{-0.1cm}
The authors would like to thank the Alexander von Humboldt (AvH) Stiftung for funding this research.

\vspace{-0.3cm}
\bibliography{bibliography}

\end{document}